\documentclass{isprs} % isprs class modified 23-04-2019 (Dennis Wittich)
\usepackage{subfigure}
\usepackage{subcaption}
\usepackage{setspace}
\usepackage{geometry} % added 27-02-2014 Markus Englich
\usepackage{epstopdf}
\usepackage{amsmath}
\usepackage[labelsep=period]{caption}  % added 14-04-2016 Markus Englich - Recommendation by Sebastian Brocks
\usepackage[british]{babel} 
\usepackage[hang]{footmisc}
\usepackage{amssymb}
\usepackage{cuted}     % preamble
\usepackage{capt-of}   % for \captionof
\usepackage{url}
\newcommand{\up}{\ensuremath{\uparrow}}
\newcommand{\down}{\ensuremath{\downarrow}}

\begin{document}

\title{Incremental Semantics-Aided Meshing from LiDAR-Inertial Odometry and RGB Direct Label Transfer}
\date{}

% KAO: Remove extra spacing

% Anonymous submissions, authors' names should not be visible
% \author{
%  Orhan Altan\textsuperscript{1}, Ian Dowman\textsuperscript{2}, Florent Lafarge\textsuperscript{3}, Clément Mallet\textsuperscript{4}, Christian Heipke\textsuperscript{5} }
\author{Muhammad Affan\textsuperscript{1}, Ville Lehtola\textsuperscript{1}, George Vosselman\textsuperscript{1}
}
% KAO: Remove extra newline
% Anonymous submissions, authors' affiliations should not be visible
\address{
	\textsuperscript{1}Faculty of Geo-Information Science and Earth Observation (ITC), 
University of Twente, 7522 NB Enschede, Netherlands\\
(m.affan, v.v.lehtola, george.vosselman)@utwente.nl
}

% If the corresponding author is NOT the final author, always add a % space before the subsequent comma, i.e.
% first author name\textsuperscript{a,}\thanks{Corresponding author} , % second author name \textsuperscript{b}, etc.
% thanks to Niclas Borlin 05-05-2016
% information on the corresponding author should not be used any longer and has been commented out
% C. Heipke, Jan 03,2024

% the use of the information of commissions and working groups should not be used any longer and has been commented out
% C. Heipke, Sept. 20,2022
%\commission{XX, }{YY} %This field is optional. If filled, XX and YY should be replaced by adequate numbers. See https://www2.isprs.org/commissions/
%\workinggroup{XX/YY} %This field is optional.
%\icwg{}   %This field is optional.

% we want to say that this method is very useful for large and complex indoor places, such as cultural buildings. applications areas, two: sparse point clouds, large point clouds. manage completeness vs. accuracy. large indoor spaces. complex large indoor environments with sparse point clouds.

% KAO: Use times symbol
\abstract{
Geometric high-fidelity mesh reconstruction from LiDAR-inertial scans remains challenging in large, complex indoor environments -- such as cultural buildings -- where point cloud sparsity, geometric drift, and fixed fusion parameters produce holes, over-smoothing, and spurious surfaces at structural boundaries. We propose a modular, incremental RGB+LiDAR pipeline that generates incremental semantics-aided high-quality meshes from indoor scans through scan frame-based direct label transfer. A vision foundation model labels each incoming RGB frame; labels are incrementally projected and fused onto a LiDAR-inertial odometry map; and an incremental semantics-aware Truncated Signed Distance Function (TSDF) fusion step produces the final mesh via marching cubes. This frame-level fusion strategy preserves the geometric fidelity of LiDAR while leveraging rich visual semantics to resolve geometric ambiguities at reconstruction boundaries caused by LiDAR point-cloud sparsity and geometric drift. We demonstrate that semantic guidance improves geometric reconstruction quality; quantitative evaluation is therefore performed using geometric metrics on the Oxford Spires dataset, while results from the NTU VIRAL dataset are analyzed qualitatively. The proposed method outperforms state-of-the-art geometric baselines ImMesh and Voxblox, demonstrating the benefit of semantics-aided fusion for geometric mesh quality. The resulting semantically labelled meshes are of value when reconstructing Universal Scene Description (USD) assets, offering a path from indoor LiDAR scanning to XR and digital modeling.
}
% uncertainties and

\maketitle
% Place this near a paragraph break; LaTeX prefers to put figure* at the top of the next page.

\begin{strip}
  \centering
  \includegraphics[width=0.96\textwidth]{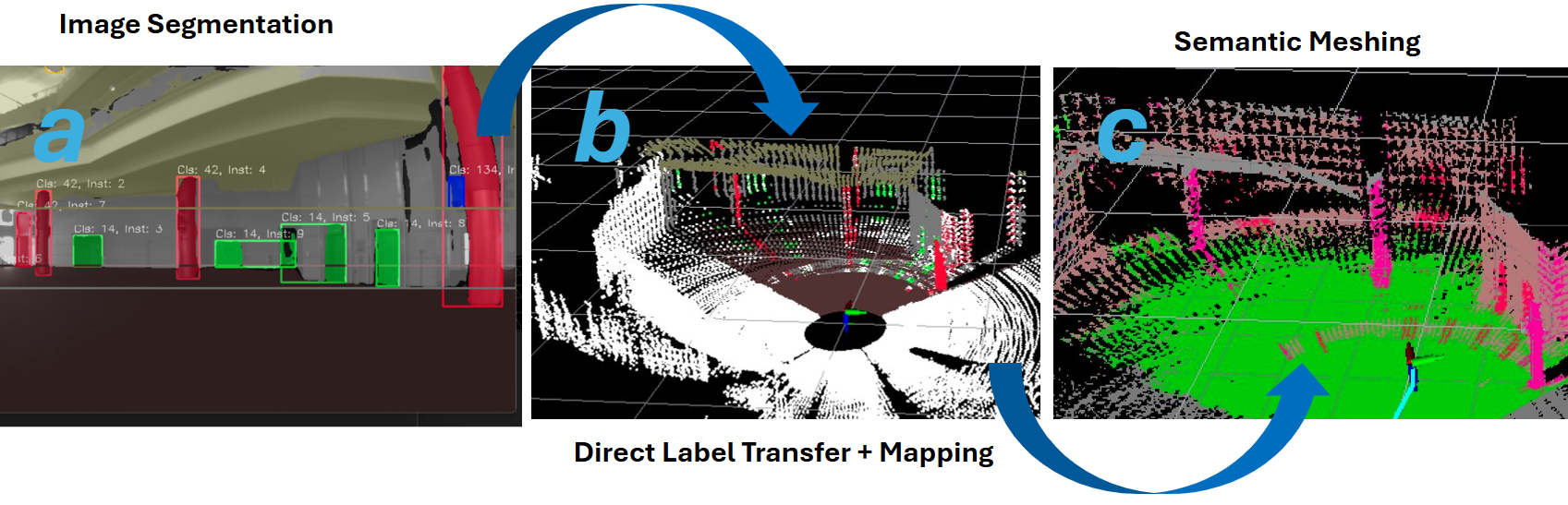}
  \captionof{figure}{Overview of the proposed pipeline: (a) direct label transfer from VFM (OneFormer) segmented RGB images onto (b) LiDAR scan frames registered via LiDAR-inertial odometry, followed by (c) semantics-aided mesh reconstruction.}
  \label{fig:teaser}
\end{strip}
%\saythanks % added 28-02-2014 Markus Englich
 
\section{Introduction}\label{introduction}
% KAO: Sloppy spacing ensures non-overfull lines. Can be removed if this is not an issue.
 \sloppy

% motivates why semantics-for-geometry matters and what we do about it
Three-dimensional reconstruction of real-world scenes is a vital enabling technology for digital twins \cite{Boje_2020}, architecture/engineering / construction workflows \cite{TANG2010829}, immersive XR content for the preservation of cultural heritage \cite{rs3061104,OpenHeritage3D2023}, and robotics simulation \cite{voxblox}. Across these domains, the common need is to convert sensor data into geometrically accurate, semantically-geometrically consistent scene models that can be analyzed, edited, and reused in downstream pipelines. Universal Scene Description (USD) has emerged as a widely adopted scene format for such interoperable 3D content \cite{pixarUSD,halacheva2025articulate3dholisticunderstanding3d}, and we refer to the conversion from raw scans to USD as Scan2USD.

Achieving high geometric fidelity in mesh reconstruction is challenging, particularly at boundaries between structurally different surfaces, where LiDAR sparsity, geometric drift, and fixed fusion parameters lead to artifacts such as holes, over-smoothing, and spurious surfaces. A promising direction is to leverage semantic scene understanding to guide the geometric reconstruction process itself: if the fusion pipeline knows that a voxel belongs to a thin railing rather than a broad wall, it can adapt its truncation distance and integration weight accordingly. Existing approaches that couple semantics with reconstruction are predominantly RGB-D-based \cite{Rosinol2021Kimera,hughes2024foundations} and assume close-range, well-textured input, while LiDAR-based meshing systems \cite{immesh,zhu2023} operate without semantic information entirely. 

In this work, we show that semantic guidance — transferred from 
RGB onto sparse LiDAR-inertial odometry maps —
improves geometric mesh reconstruction. Our contributions are:

% In this work, we show that semantic guidance — transferred from 
% RGB onto sparse LiDAR-inertial odometry maps — measurably 
% improves geometric mesh reconstruction in these challenging 
% settings.

% Our central claim is that semantic guidance improves the quality of geometric reconstruction. We make the following contributions:
\begin{itemize}
    \item A modular, incremental pipeline that transfers Vision Foundation Model (VFM)-based panoptic labels from RGB images to sparse LiDAR-inertial odometry point clouds through ego-motion-compensated projection. Multi-stage boundary filtering improves RGB–LiDAR correspondence under sparse sampling and cross-modal temporal misalignment, in contrast to RGB-D systems that typically rely on dense, time-synchronous 2D–3D associations. The modular design makes the pipeline compatible with other segmentation backbones and readily extensible to multi-camera configurations.
    \item A label-aware TSDF integration scheme — combining Dirichlet-inspired per-voxel semantic fusion, class-conditioned dynamic truncation, and temporal label consolidation — that adapts fusion to class-specific geometric characteristics, improving mesh quality over geometry-only baselines.
    \item Evaluation on the Oxford Spires and NTU VIRAL datasets demonstrating improvements over ImMesh and Voxblox, with per-voxel uncertainty metrics that separately characterize semantic and geometric inconsistencies.
\end{itemize}
The remainder of this paper is organized as follows. Section 2 reviews related work on volumetric reconstruction, LiDAR-based meshing, and semantic 3D mapping. Section 3 describes the proposed method, including the label transfer mechanism, label-aware TSDF fusion, and mesh extraction. Section 4 presents the uncertainty analysis framework. The results are reported in Section 5, and Section 6 concludes the paper.

\section{Related Work}
\label{sec:related_work}
% builds the scientific basis and provides the gap statement at the end of 2.3
\subsection{Volumetric 3D Reconstruction}
Volumetric approaches based on Truncated Signed Distance Fields (TSDFs) have become a standard framework for dense 3D reconstruction. KinectFusion \cite{kinnectfusion} demonstrated real-time volumetric fusion from RGB-D streams, and subsequent systems such as Kintinuous \cite{Whelan2013Kintinuous} and InfiniTAM \cite{InfiniTAM_ISMAR_2015} extended this to larger workspaces. Scalability was further addressed through voxel hashing \cite{10.1145/2508363.2508374}, which allocates memory only for occupied regions, enabling reconstruction of large scenes without a fixed grid. BundleFusion \cite{Dai2017BundleFusion} added global bundle adjustment for drift-free reconstruction, while Voxblox \cite{voxblox} introduced a modular TSDF and ESDF integration pipeline with online mesh extraction that has influenced many later systems \cite{reijgwart2020voxgraph,grinvald2019volumetric}. More recently, adaptive-resolution TSDF schemes \cite{10601297} and neural implicit representations \cite{zhu2023} have been explored as alternatives to uniform voxel grids. A common limitation of these systems is that they treat geometry and semantics independently, if at all — fusion parameters such as truncation distance and integration weight are fixed globally, regardless of the surface type being reconstructed.

\subsection{LiDAR-Based Meshing}

% In contrast to RGB-D, LiDAR SLAM traditionally produces point clouds or sparse surfel maps. The work by \cite{immesh} introduced ImMesh, a real-time LiDAR SLAM and meshing framework. ImMesh maintains a sparse voxel map and “incrementally reconstructs the triangle mesh on the fly” for each new LiDAR scan \cite{immesh}. It works by projecting points within each occupied voxel to a local 2D plane and performs per-voxel meshing (pull–commit–push) to grow the mesh iteratively. This voxel-wise approach is optimized for CPU efficiency, and claimed to be the first work in the literature to reconstruct online the triangle mesh of large-scale scenes on a standard CPU. Similarly, \cite{zhu2023} propose Mesh-LOAM, extending the popular LiDAR Odometry and Mapping (LOAM) paradigm with mesh output. Mesh-LOAM maintains a voxel grid for the scene and updates it with each scan; it then fits a local triangular mesh over occupied voxels and uses a “point-to-mesh” odometry against this mesh. In practice, the system incrementally updates voxels and applies Marching Cubes locally to generate faces, enabling real-time pose estimation and a compact mesh map.
While the volumetric methods above predominantly target RGB-D input, a separate line of work has pursued mesh reconstruction directly from LiDAR scans. ImMesh \cite{immesh} maintains a sparse voxel map and performs incremental per-voxel triangulation via a pull–commit–push scheme, claimed to be the first system to reconstruct triangle meshes of large-scale scenes online on a standard CPU. Mesh-LOAM \cite{zhu2023} extends the LOAM paradigm by maintaining a voxel grid updated per scan and applying local marching cubes to generate mesh faces, enabling joint pose estimation and mesh mapping. Both systems demonstrate real-time capability and produce geometrically detailed meshes. However, they are purely geometric: fusion parameters are class-agnostic, and no semantic information is carried on the output mesh. ImMesh and Voxblox serve as geometry-only baselines in our evaluation.

\subsection{Semantic 3D Reconstruction}
% Meshing pipelines strive to produce highly fidelity, semantically-understandable, textured 3D scenes for AR/VR or simulation. Although relatively few recent works focus on semantic meshing, there is growing interest in embedding scene understanding into the reconstruction. In practice, meshes and geometry are required to have high-level formats for AR/VR compatibility, one of those is the Pixar Universal Scene Description (USD) format \cite{pixarUSD}. USD provides a rich scenegraph structure to pack meshes, materials, and semantic groupings together, making it popular for AR/VR/XR pipelines. Recent research has increasingly explored Universal Scene Description (USD) as a target output format for 3D scanning pipelines, enabling structured, semantically rich scene representations that can be imported directly into graphics engines such as Unreal Engine or Apple’s RealityKit \cite{halacheva2025articulate3dholisticunderstanding3d,hughes2024foundations}. The convergence of scene understanding, scene representation, and utilization suggests a future where 3D scanning not only captures geometry, but semantically contextualizes and structures it for downstream AR/VR and robotics applications.
A growing body of work integrates 2D semantic predictions into the 3D reconstruction loop. SemanticFusion \cite{McCormac2017SemanticFusion} fuses CNN-based class predictions into an ElasticFusion surfel map, providing online per-surfel semantics. Kimera \cite{Rosinol2021Kimera} builds a metric–semantic SLAM stack with TSDF meshing on a factor-graph backend, while its successor Hydra \cite{hughes2024foundations} advances this to a layered volumetric scene graph encoding geometry, semantics, and instances in real time. At the instance level, PanopticFusion \cite{Nakajima2019PanopticFusion} couples Mask R-CNN with a TSDF backend to maintain per-instance volumes from RGB-D streams; MaskFusion \cite{Runz2018MaskFusion} leverages instance masks with surfel fusion for real-time object tracking; and Fusion++ \cite{McCormac2018FusionPP} builds per-object TSDF submaps from 2D instance masks.
A unifying characteristic of these systems is their reliance on RGB-D input: the co-located depth map provides a dense, per-pixel association between the 2D segmentation and 3D geometry. This tight coupling simplifies label transfer but restricts applicability to close-range scenarios (typically $<$ 10 m) with well-textured surfaces. In large indoor environments — museums, stations, cultural buildings — distances routinely exceed this range, lighting varies widely, and surfaces may be textureless or specular. LiDAR sensors are better suited to these conditions, yet to our knowledge no existing system transfers VFM-derived semantic labels onto a LiDAR-inertial odometry map and exploits them to improve the geometric TSDF fusion process. Our work addresses this gap.

% \subsection{Evaluation Metrics for 3D Reconstruction}
% Geometric reconstruction quality is commonly assessed using point-to-point or point-to-surface distance metrics between a reconstructed mesh and a ground-truth reference. Standard measures include accuracy (mean or RMSE distance from reconstruction to ground truth), completeness (mean or RMSE distance from ground truth to reconstruction), and their harmonic combination as an F1 score at a given distance threshold \cite{Schoeps2019BADSLAM, Dai2017BundleFusion}. BundleFusion \cite{Dai2017BundleFusion} established this protocol for RGB-D reconstruction, and subsequent LiDAR-based systems such as ImMesh \cite{immesh} and Mesh-LOAM \cite{zhu2023} adopt the same framework, typically sampling reconstructed meshes to a fixed point count and aligning to ground truth via ICP before computing distances. We follow this established protocol in our evaluation Section \ref{sec:Results}.

\section{Methods}\label{sec:methods}

\begin{figure}[ht!]
  \centering
  \includegraphics[width=\linewidth]{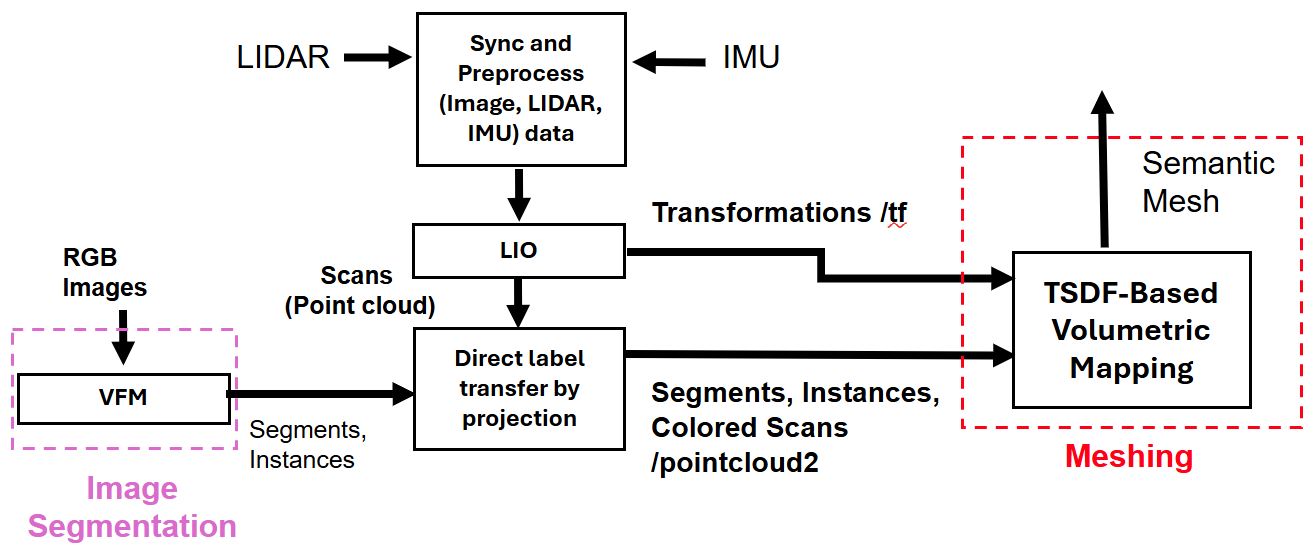}
  \caption{The flowchart of the proposed method.}
  \label{fig:method}
\end{figure}

Our system (Figure~\ref{fig:method}) incrementally produces a semantically labelled triangle mesh from synchronized LiDAR, RGB, and IMU streams acquired from a calibrated multi-sensor platform, exportable as a USD asset (Section~\ref{sec:usd_export}) for interoperable use in XR and digital modeling pipelines. The pipeline operates at LiDAR rate in three stages: (1)~panoptic segmentation of RGB frames using OneFormer~\cite{Jain_2023_CVPR} (dotted pink box in Figure~\ref{fig:method}); (2)~tightly coupled LiDAR-inertial odometry via FAST-LIO2, with direct label transfer from image masks onto deskewed point clouds (Section~\ref{sec:label_transfer}); and (3)~semantics-aware TSDF fusion and marching cubes meshing, where truncation and weighting adapt per voxel based on fused labels (Sections~\ref{sec:label_aware_fusion}--\ref{sec:mesh_extraction}, dotted red box in Figure~\ref{fig:method}). The following subsections detail each stage.

% ============================================================
\subsection{LiDAR-Inertial Odometry}\label{sec:lio}
% ============================================================
For LiDAR-inertial odometry (LIO), we adopt FAST-LIO2~\cite{fastlio2} as the localization backbone. Unlike classical feature-based LiDAR odometry (e.g., LOAM), which extracts planar/edge features and can be bottlenecked by feature selection, FAST-LIO2 performs tightly coupled fusion of raw LiDAR and IMU measurements and exploits an incremental k-d tree (ikd-Tree) for fast nearest-neighbor operations. This allows registration of tens of thousands of points per scan in real time on CPU, providing low-drift pose estimates and a reliable sparse map onto which our volumetric fusion and semantic meshing are built.
 
We use LIO instead of full SLAM with loop closure to avoid discontinuous pose corrections that would invalidate the incrementally built TSDF and require costly re-integration. For the indoor trajectories evaluated (several hundred meters), tightly coupled LIO drift remains within the voxel resolution tolerance.
 
% ============================================================
\subsection{Direct Label Transfer}\label{sec:label_transfer}
% ============================================================
The direct label transfer module projects 2D panoptic segmentation onto 3D LiDAR points using OneFormer~\cite{Jain_2023_CVPR}. For each scan, the IMU state is forward-propagated to the camera timestamp~$t_C$, yielding a dynamic LiDAR-to-camera transform
\begin{equation}
  \mathbf{T}_{CL}
  = \bigl(\mathbf{T}_{WI}(t_C) \cdot \mathbf{T}_{IL} \cdot \mathbf{T}_{LC}\bigr)^{-1}
    \cdot \mathbf{T}_{WI}(t_L) \cdot \mathbf{T}_{IL},
  \label{eq:T_CL}
\end{equation}
where $\mathbf{T}_{WI}(t)$ is the world-to-IMU pose from the LIO filter, $t_L$ the LiDAR timestamp, and $\mathbf{T}_{IL}$, $\mathbf{T}_{LC}$ are the static IMU-to-LiDAR and LiDAR-to-camera extrinsics. Each deskewed point is projected via a pinhole-plus-distortion model with per-point motion correction from the instantaneous camera velocity. To suppress label bleed, morphological mask erosion, boundary distance rejection, and depth-discontinuity checks filter unreliable projections; each surviving point receives the label of the highest-scoring mask at its projected location, while ambiguous and out-of-field-of-view points enter the TSDF as geometry-only observations.

% ============================================================
\subsection{Label-Aware TSDF Fusion}\label{sec:label_aware_fusion}
% ============================================================
After direct label transfer, each LiDAR scan arrives at the TSDF integrator as a mixture of semantically labelled and unlabelled points, together with the LIO-estimated sensor pose. These are integrated into a voxel grid using a modified TSDF scheme that maintains per-voxel semantic state alongside the signed distance and weight fields. The key modifications over standard TSDF integration (e.g., Voxblox) are described below.
\paragraph{Per-voxel semantic fusion.}
Each voxel~$v$ maintains a small label histogram $H_v(\ell)$ and a current label~$\ell_v$.
During integration, all labelled points that map to~$v$ in the current pass are bundled and a provisional label~$\tilde{\ell}_v$ is chosen by majority vote among their non-zero labels. We then add a \emph{single} confidence increment to the histogram entry of~$\tilde{\ell}_v$:
\[
  H_v(\tilde{\ell}_v) \leftarrow H_v(\tilde{\ell}_v) + \eta(r),
\]
where $r = \lVert \mathbf{p}_L \rVert_2$ is the range of a representative LiDAR return in the sensor frame. The confidence term~$\eta(r)$ follows a log-normal decay model:
\begin{equation}
  \eta(r)=
  \mathrm{LogNormal}\!\big(\ln(\max\{0,\,r-r_0\});\,\mu,\sigma\big),
  \label{eq:lognormal-weight}
\end{equation}
with short-range offset~$r_0$ and log-normal parameters~$(\mu,\sigma)$. This encodes the observation that label reliability degrades with range due to increasing projection uncertainty and decreasing LiDAR angular resolution. After the update, the voxel label is set to $\ell_v \leftarrow \arg\max_{\ell} H_v(\ell)$. To avoid semantic erosion, an existing non-zero label is never overwritten by background.
\paragraph{Greedy segment--label assignment.}
Before voxel updates, incoming points are grouped into segments $s \in \mathcal{S}$ and each segment is cast into the current map, tallying its overlap with existing map labels using voxels within a narrow TSDF band. We greedily assign~$s$ to the label that maximizes this overlap, enforcing per-frame uniqueness; segments with no plausible match receive a new map label. This reduces label fragmentation under sparse coverage.
\paragraph{Temporal label consolidation.}
Across frames, we maintain pairwise label co-occurrence counts $\Pi(\ell_a, \ell_b)$. When~$\Pi$ exceeds a threshold, the older label is merged into the newer one and per-voxel histograms containing the old label are rewritten in place. This yields stable, long-lived identities as evidence accumulates.
% ============================================================
\subsection{Dynamic Label-Aware Truncation}
\label{sec:dynamic_truncation}
% ===========================================================
Standard TSDF integration applies a fixed truncation distance~$\mu_0$ to all voxels, compromising between thin structures and large planar surfaces. Semantic labels provide a prior on expected geometry: \emph{railing} voxels use tighter truncation to preserve fine detail, while \emph{wall} or \emph{floor} voxels use wider truncation for completeness and noise averaging.
 
We define a dynamic truncation distance that adapts to both the semantic label and the LiDAR range:
\begin{equation}
  \mu(\hat{\ell}, r)
  = \mathrm{clip}\!\Big(
    \mu_0 \, f_{\mathrm{label}}(\hat{\ell}) \, f_{\mathrm{range}}(r),\;
    \mu_{\min},\,\mu_{\max}
  \Big),
  \label{eq:dyn_mu}
\end{equation}
where $\hat{\ell}$ is the majority-voted voxel label, $f_{\text{label}}(\cdot)$ is a class-specific multiplier derived from geometric priors on expected surface thickness (e.g., $f_{\text{label}}(\text{railing}) = 0.5$, $f_{\text{label}}(\text{wall}) = 1.2$), $f_{\text{range}}(r)$ is a bucketed range multiplier compensating for increased range noise, and $\text{clip}(\cdot, \mu_{\min}, \mu_{\max})$ enforces stability limits. Class multipliers are set empirically.
 
The TSDF update incorporates range-dependent weighting and sparsity compensation. Given sensor origin~$\mathbf{o}$, world-frame hit point~$\mathbf{p}_G$, and voxel centre~$\mathbf{x}$, the signed distance is $\phi = \phi(\mathbf{o}, \mathbf{p}_G, \mathbf{x})$ (positive in front of the surface). The base weight encodes sensor confidence:
\[
  w_{\mathrm{base}}(\mathbf{p}_L) =
  \begin{cases}
    1, & \text{(constant weighting)},\\[4pt]
    \dfrac{1}{z^2}, & \text{otherwise, with } z = \bigl|(\mathbf{p}_L)_z\bigr|,
  \end{cases}
\]
where the inverse-square mode down-weights distant points that carry higher range noise. Using this base weight $w = w_{\mathrm{base}}(\mathbf{p}_L)$ and a dynamic truncation band $\mu = \mu(\hat{\ell}, r)$, the TSDF voxel with current value/weight $(\phi_v, w_v)$ is updated as:
\begin{subequations}\label{eq:tsdf_update}
\begin{align}
  \tilde{w} &\leftarrow
  \begin{cases}
    w \dfrac{\mu+\phi}{\mu-\varepsilon}, & \phi<-\varepsilon \;\text{(behind-surface dropoff)},\\[4pt]
    w, & \text{otherwise},
  \end{cases}
  \label{eq:tsdf_drop}\\[4pt]
  \tilde{w} &\leftarrow
  \begin{cases}
    \tilde{w}\,\gamma, & |\phi|<\mu \;\text{(sparsity compensation)},\\
    \tilde{w}, & \text{otherwise},
  \end{cases}
  \label{eq:tsdf_sparse}\\[4pt]
  w_{\mathrm{new}} &\leftarrow \min\bigl(w_{\max},\, w_v+\tilde{w}\bigr),
  \label{eq:tsdf_wnew}\\[4pt]
  \phi_{\mathrm{new}} &\leftarrow \dfrac{\tilde{w}\,\phi + w_v\,\phi_v}{w_{\mathrm{new}}},
  \label{eq:tsdf_phinew}\\[4pt]
  \phi_v &\leftarrow \mathrm{clip}(\phi_{\mathrm{new}},-\mu,\mu),\quad
  w_v \leftarrow w_{\mathrm{new}}.
  \label{eq:tsdf_assign}
\end{align}
\end{subequations}
Here $\varepsilon \approx \Delta$ (equal to the voxel size, 0.1\,m) attenuates weight behind the surface to suppress back-face bleed, $\gamma = 1.5$ is a sparsity compensation factor that counteracts the inherently lower point density of LiDAR returns, and $w_{\max} = 255$ clamps the weight to prevent unbounded accumulation. All three values were set empirically.

\subsection{Mesh Extraction and Label Assignment}\label{sec:mesh_extraction}
After each integration step, the mesh is extracted via marching cubes on the TSDF zero-level set. Each triangle inherits the semantic label $\ell_v = \arg\max_\ell H_v(\ell)$ and instance identifier from the nearest voxel centre.

% ============================================================
\subsection{USD Export}\label{sec:usd_export}
% ============================================================
The semantically labelled mesh can be exported as a USD asset by grouping triangles by instance identifier into \texttt{UsdGeom.Mesh} primitives with class labels as metadata, establishing a direct path from LiDAR-inertial scanning to USD-compatible engines such as Unreal Engine or RealityKit.

\section{Uncertainty Analysis}\label{sec:section4}
Standard metrics such as accuracy, completeness, and F1 (Section~\ref{sec:Results}) quantify global reconstruction quality but do not decompose error sources. In our multi-modal pipeline, inconsistencies originate from two distinct sources: geometric (LIO drift, range noise, TSDF discretisation) and semantic (segmentation errors, label flickering, projection misalignment). Since our central claim is that semantic guidance improves geometric reconstruction, we maintain per-voxel uncertainty scores that separately characterise geometric and semantic evidence. This decomposition reveals whether inconsistencies are attributable to the geometric backend, the semantic frontend, or their interaction.

\subsection{Notation and Inputs}
We maintain per-TSDF-voxel semantic and geometric 
\emph{uncertainty scores} $s_{\text{sem}}, s_{\text{geom}} \in [0,1]$, 
where higher values indicate greater uncertainty about 
reconstruction quality at that voxel. At each integration step, we compute per-voxel evidence cues $\{e_{\text{sem}}, e_{\text{geom}}\}$ from local observations; throughout this section, all $e \in [0,1]$ are oriented such that $e = 0$ denotes full agreement and $e \to 1$ denotes maximal 
disagreement within the respective modality. These cues are 
fused into the persistent scores via an exponential moving 
average (EMA).

Let \(\phi(\mathbf{x})\) denote the TSDF grid, and let \(v\) index a voxel
with center \(\mathbf{x}_v\) and voxel size \(\Delta\). From the label voxel, we have the winning label $\ell(\mathbf{v})$ and a small histogram of label confidences. We denote by \(\mathcal{N}_6(v)\) the 6-neighborhood of \(v\) in the voxel grid.

\subsection{Semantic Uncertainty Evidence}
\paragraph{Dirichlet-inspired Per-voxel Consensus}
We use the voxel-level label histogram \(H_v(\ell)\ge 0\) as unnormalized evidence analogous to a Dirichlet distribution's concentration parameters over semantic labels (excluding background).  Define
\[
S_v \;=\; \sum_{\ell\neq 0} H_v(\ell), 
\qquad 
M_v \;=\; \max_{\ell\neq 0} H_v(\ell).
\]
The \emph{label consensus} over each \(v\) is the dominance of the most frequent non-background label:
\[
\mathrm{consensus}(v) \;=\;
\begin{cases}
1, & \text{if } S_v=0,\\[2pt]
\dfrac{M_v}{S_v}, & \text{otherwise.}
\end{cases}
\]
Thus, \(\mathrm{consensus}(v)\in[0,1]\), where \(1\) implies evidence supporting a single non-background label, and lower values reflect mixed evidence or diffuse support, computed as:
\[
e_{\text{conc}}\;=\; (1.0 - {consensus}(v) );
\]
In the above formulation, $e_{\text{conc}}$ is zero under full label consensus and increases toward one as the distribution becomes more mixed, providing a simple Dirichlet-entropy–like impurity measure without explicit histogram normalization. 

\paragraph{Neighbor Disagreement.}
We test for any nonzero 6-neighbor with a different label:
\begin{equation}
  e_{\text{neigh}} \;=\; 
  \begin{cases}
    1, & \exists\, u\!\in\!\mathcal{N}_6(v):\;
          \ell(u)\neq 0,\;\ell(u)\neq \ell(v),\\[2pt]
    0, & \text{otherwise.}
  \end{cases}
\end{equation}

\paragraph{Semantic fusion.}
With nonnegative weights $w_{\text{conc}},w_{\text{neigh}}$, we form
\begin{equation}
  e_{\text{sem}} \;=\;
  \frac{w_{\text{conc}}\,e_{\text{conc}}
      + w_{\text{neigh}}\,e_{\text{neigh}}}
       {w_{\text{conc}}+w_{\text{neigh}}}.
  \label{eq:esem}
\end{equation}

\subsection{Geometric Uncertainty Evidence}
We combine three cues: TSDF gradient excess, curvature from points,
and normal-change across neighbors.

\paragraph{TSDF gradient (excess over 1)}
The Voxblox integrator maintains a TSDF \(\phi(\mathbf{x})\) in \emph{world units} (meters), which ideally approximates a signed distance function near surfaces. For a voxel \(v\) with center \(\mathbf{x}_v\) and grid spacing \(\Delta\), we approximate the spatial derivatives of \(\phi\) by central differences:
\begin{subequations}\label{eq:grad_vox}
\begin{align}
\partial_x \phi(\mathbf{x}_v) &\approx 
\frac{\phi(\mathbf{x}_v+\Delta\,\mathbf{e}_x)-\phi(\mathbf{x}_v-\Delta\,\mathbf{e}_x)}{2\Delta},\\
\partial_y \phi(\mathbf{x}_v) &\approx 
\frac{\phi(\mathbf{x}_v+\Delta\,\mathbf{e}_y)-\phi(\mathbf{x}_v-\Delta\,\mathbf{e}_y)}{2\Delta},\\
\partial_z \phi(\mathbf{x}_v) &\approx 
\frac{\phi(\mathbf{x}_v+\Delta\,\mathbf{e}_z)-\phi(\mathbf{x}_v-\Delta\,\mathbf{e}_z)}{2\Delta},
\end{align}
\end{subequations}
where \(\mathbf{e}_x,\mathbf{e}_y,\mathbf{e}_z\) are the canonical unit vectors in \(\mathbb{R}^3\).
The gradient magnitude in world units is then
\[
\|\nabla \phi(\mathbf{x}_v)\| \;\approx\;
\sqrt{
\bigl(\partial_x \phi(\mathbf{x}_v)\bigr)^2 +
\bigl(\partial_y \phi(\mathbf{x}_v)\bigr)^2 +
\bigl(\partial_z \phi(\mathbf{x}_v)\bigr)^2
}.
\]
For a smooth Euclidean signed distance field one expects \(\|\nabla\phi(\mathbf{x})\|\approx 1\) near the zero level set. We therefore measure \emph{excess} gradient magnitude above this ideal value:
\begin{equation}
  e_{\text{grad}}(v) \;=\;
  \mathrm{clip}_{[0,1]}
  \!\left(\frac{\|\nabla\phi(\mathbf{x}_v)\| - 1}{\tau_{\text{grad}}}\right),
  \label{eq:egrad}
\end{equation}
with scale \(\tau_{\text{grad}}>0\) controlling sensitivity. Large values of \(e_{\text{grad}}\) indicate strong deviations from a distance-like TSDF (e.g.\ due to fusion inconsistencies, discretization, or noisy observations).

\paragraph{Curvature from raw points}
From a voxel-neighborhood point set $\{\mathbf{x}_i\}$ we compute the
covariance $\mathbf{C}$ in the global (world) frame and use the
$\lambda_{\min}/\mathrm{tr}(\mathbf{C})$ ratio, where $\lambda_{\min}(\mathbf{C})$ is the smallest eigenvalue of~$\mathbf{C}$ and $\mathrm{tr}(\mathbf{C})$ is its trace:
\begin{equation}
  e_{\text{curv}} \;=\; \frac{\lambda_{\min}(\mathbf{C})}{\mathrm{tr}(\mathbf{C})}
  \;\in\;[0,1].
  \label{eq:ecurv}
\end{equation}

\paragraph{Normal-change (TSDF)}
Let \(\mathbf{n}(u)\) be the normalized TSDF gradient at voxel
\(u\), i.e.\ \(\mathbf{n}(u) = \nabla \phi(\mathbf{x}_u) / \|\nabla \phi(\mathbf{x}_u)\|\) whenever the gradient is nonzero. For a center voxel \(v\) we average the angle (mapped to \([0,1]\)) between \(\mathbf{n}(v)\) and the normals of its 6-neighbors:
\begin{align}
  \theta(u) &= \arccos\big(|\mathbf{n}(v)^\top \mathbf{n}(u)|\big),\\
  e_{\text{nchg}}(v) &= \frac{1}{|\mathcal{S}|}\sum_{u\in\mathcal{S}}
  \min\!\Bigl(1,\frac{\theta(u)}{\pi/2}\Bigr),
  \quad \mathcal{S}=\mathcal{N}_6(v).
  \label{eq:enchg}
\end{align}

\paragraph{Surface-band gating}
Let \(d=\lvert\phi(v)\rvert\) be the signed-distance magnitude at voxel \(v\), and let \(b=\kappa\,\Delta\) denote the surface-band width, where \(\Delta\) is the voxel size and \(\kappa>0\) is a \emph{global} bandwidth parameter (typically $\kappa = 2$--$4$).
We softly gate two geometric evidence terms,
\(e_{\text{curv}}\) and \(e_{\text{nchg}}\). Define
\[
  g(d) \;=\; \operatorname{clip}_{[\alpha,\,1]}\!\Bigl(1-\tfrac{d}{b}\Bigr),
\]
and update
\[
  e_{\text{curv}} \leftarrow g(d)\,e_{\text{curv}}, 
  \qquad
  e_{\text{nchg}} \leftarrow g(d)\,e_{\text{nchg}},
\]
where \(\alpha\in[0,1]\) is a \emph{global} attenuation floor that sets the minimum off-surface weighting.
Thus, \(\kappa\) controls how far from the zero level set geometric evidence contributes, while \(\alpha\) limits how much that evidence is down-weighted away from the surface.
If soft gating is disabled, geometric cues are set to zero whenever \(d>b\).

\paragraph{Geometric fusion}
All geometric cues are fused in a way that the gradient cue is masked to contribute only in geometrically
interesting regions,
\begin{equation}
  e_{\text{grad}} \leftarrow \max\!\big(e_{\text{curv}},\, e_{\text{nchg}}\big)\cdot e_{\text{grad}},
\end{equation}
then fuse with nonnegative weights
$w_{\text{curv}},w_{\text{nchg}},w_{\text{grad}}$:
\begin{equation}
  e_{\text{geom}} \;=\;
  \frac{w_{\text{curv}}\,e_{\text{curv}}
      + w_{\text{nchg}}\,e_{\text{nchg}}
      + w_{\text{grad}}\,e_{\text{grad}}}
       {w_{\text{curv}}+w_{\text{nchg}}+w_{\text{grad}}}.
  \label{eq:egeom}
\end{equation}

% \subsubsection{EMA Fusion Over Time \& Thresholding}
% For each voxel we maintain persistent boundary uncertainty scores for the semantic and geometric evidence. These scores are updated over time using an exponential moving average (EMA) with smoothing factor $\beta\!\in\!(0,1]$:
% \begin{align}
%   s_{\mathrm{sem}}  &\leftarrow (1-\beta)\,s_{\mathrm{sem}}  + \beta\,e_{\mathrm{sem}}, \\
%   s_{\mathrm{geom}} &\leftarrow (1-\beta)\,s_{\mathrm{geom}} + \beta\,e_{\mathrm{geom}}.
% \end{align}

% A hard decision is then obtained by thresholding the temporally smoothed scores:
% \begin{equation}
%   \mathrm{is\_sem}=\big[s_{\mathrm{sem}}\ge \tau_{\mathrm{sem}}\big], \qquad
%   \mathrm{is\_geom}=\big[s_{\mathrm{geom}}\ge \tau_{\mathrm{geom}}\big],
% \end{equation}
% where $\tau_{\mathrm{sem}}$ and $\tau_{\mathrm{geom}}$ are semantic and geometric decision thresholds respectively.
\subsubsection{EMA Fusion Over Time \& Thresholding}
The persistent uncertainty scores are updated via an 
exponential moving average (EMA) with smoothing factor 
$\beta \in (0,1]$, in which each past observation's influence 
decays by $(1-\beta)$ per integration step:
\begin{align}
  s_{\mathrm{sem}}  &\leftarrow (1-\beta)\,s_{\mathrm{sem}}  
    + \beta\,e_{\mathrm{sem}}, \\
  s_{\mathrm{geom}} &\leftarrow (1-\beta)\,s_{\mathrm{geom}} 
    + \beta\,e_{\mathrm{geom}}.
\end{align}
Binary flags are obtained by thresholding:
\begin{equation}
  \mathrm{is\_sem}\!=\![s_{\mathrm{sem}}\!\ge\!\tau_{\mathrm{sem}}],\quad
  \mathrm{is\_geom}\!=\![s_{\mathrm{geom}}\!\ge\!\tau_{\mathrm{geom}}],
\end{equation}
where $[\cdot]$ denotes the Iverson bracket (1 if the condition 
holds, 0 otherwise) and $\tau_{\mathrm{sem}}$, 
$\tau_{\mathrm{geom}}$ are decision thresholds. These flags partition voxels into four diagnostic categories---semantic uncertainty only, geometric uncertainty only, both, or neither---thereby enabling reconstruction errors to be decomposed by source (Section~\ref{sec:UncertainityResults}).

\section{Results}\label{sec:Results}
\subsection{Qualitative Results}\label{sec:Qualitative}
\subsubsection{Oxford Spires Scene.}
\begin{figure*}[t]
    \centering
    \subfigure[ImMesh]{\includegraphics[width=0.30\textwidth]{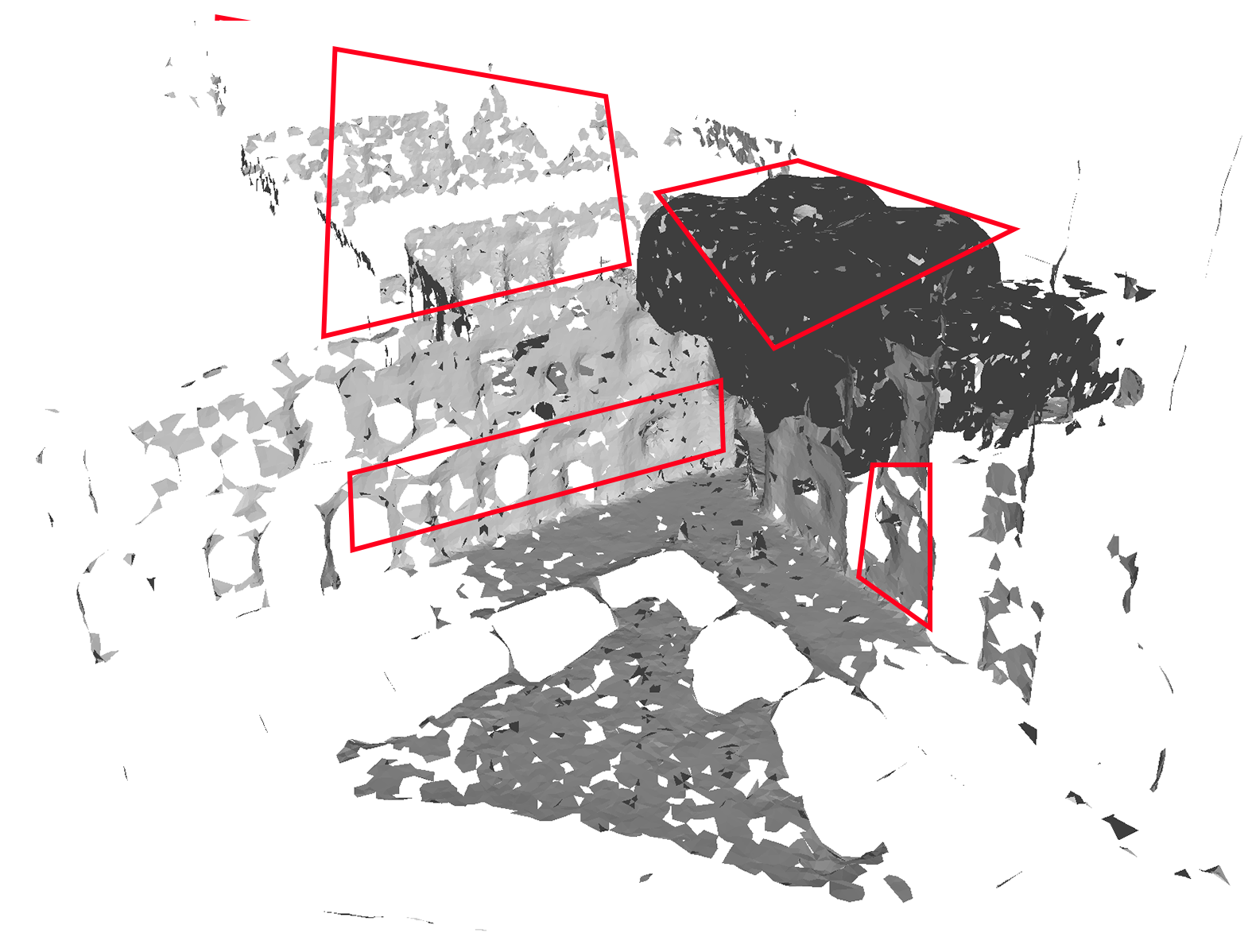}}
    \label{ImMesh}
    \hspace{0.1in} % Optional: adds horizontal space between subfigures
    \subfigure[Voxblox]{\includegraphics[width=0.30\textwidth]{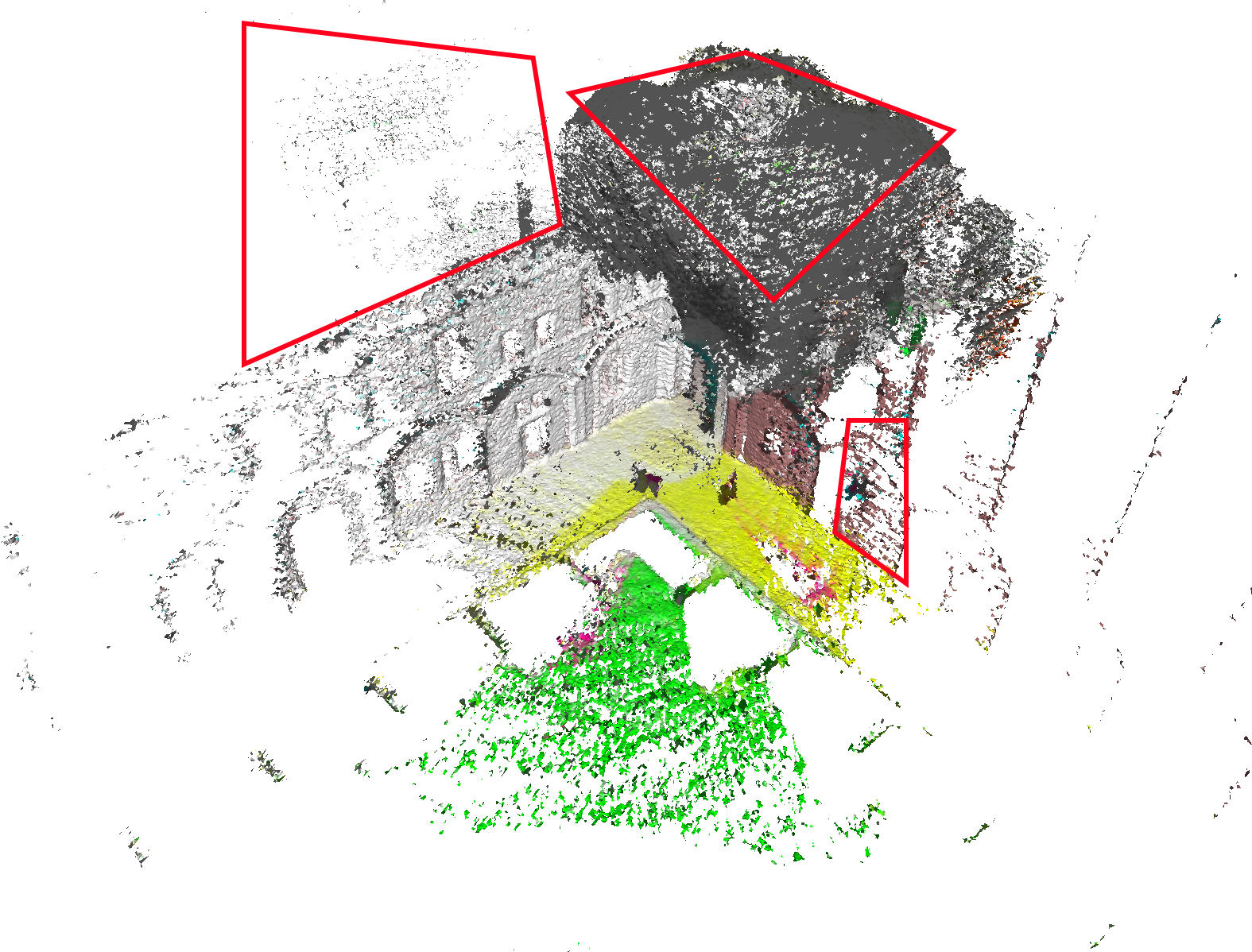}}
    \hspace{0.1in} % Optional: adds horizontal space between subfigures
    \subfigure[Ours]
    {\includegraphics[width=0.30\textwidth]{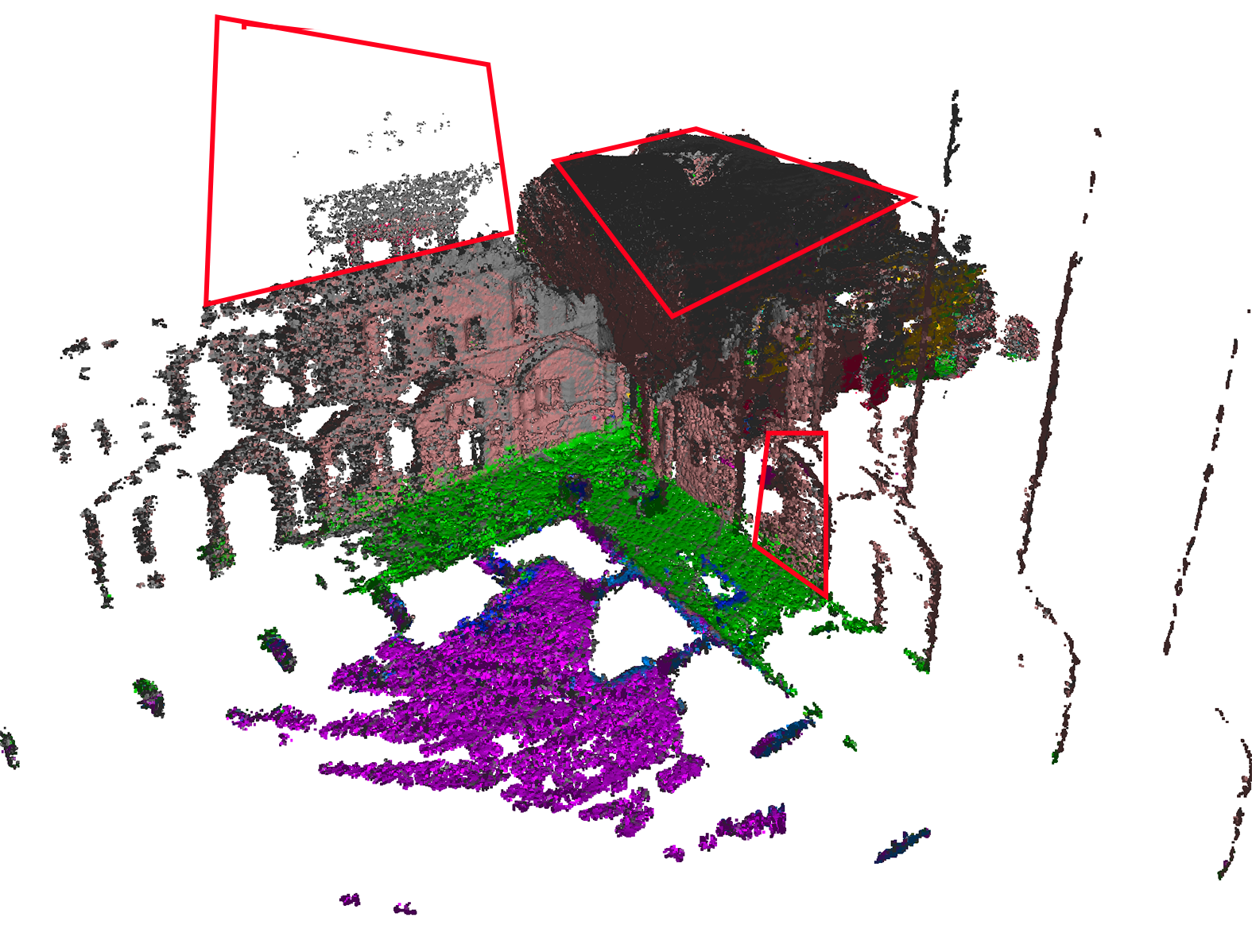}}
    \caption{3D reconstruction of Oxford Spires (Christ Church College Scene) dataset}
    \label{fig:qualitative comparison oxford-spires}
\end{figure*}

In Figure~\ref{fig:qualitative comparison oxford-spires}c, our reconstruction exhibits robust adaptability to both outdoor and indoor spaces. The generated mesh recovers both dense ground structures and tall architectural features, such as spires and arches, with consistent color integration balancing smoothness and completeness.
Voxblox (Figure~\ref{fig:qualitative comparison oxford-spires}b) again suffers from faded texture and incomplete facade geometry, while ImMesh (Figure~\ref{fig:qualitative comparison oxford-spires}a) reconstructs coarse walls but omits many thin high-frequency structures.
\subsubsection{NTU VIRAL Scene.}
\begin{figure*}[t]
    \centering
    \subfigure[ImMesh]{\includegraphics[width=0.30\textwidth]{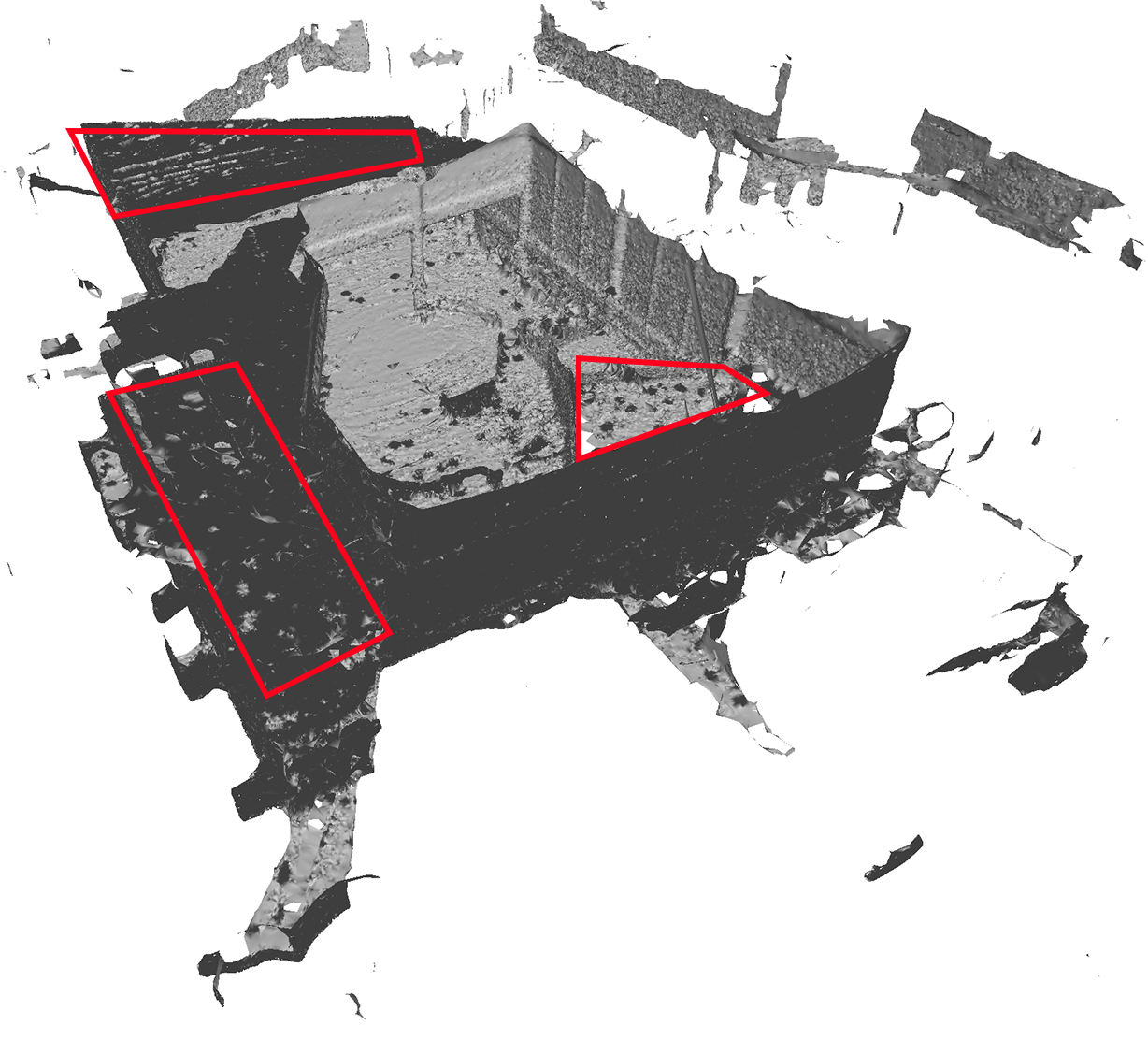}}
    \hspace{0.1in} % Optional: adds horizontal space between subfigures
    \subfigure[Voxblox]{\includegraphics[width=0.30\textwidth]{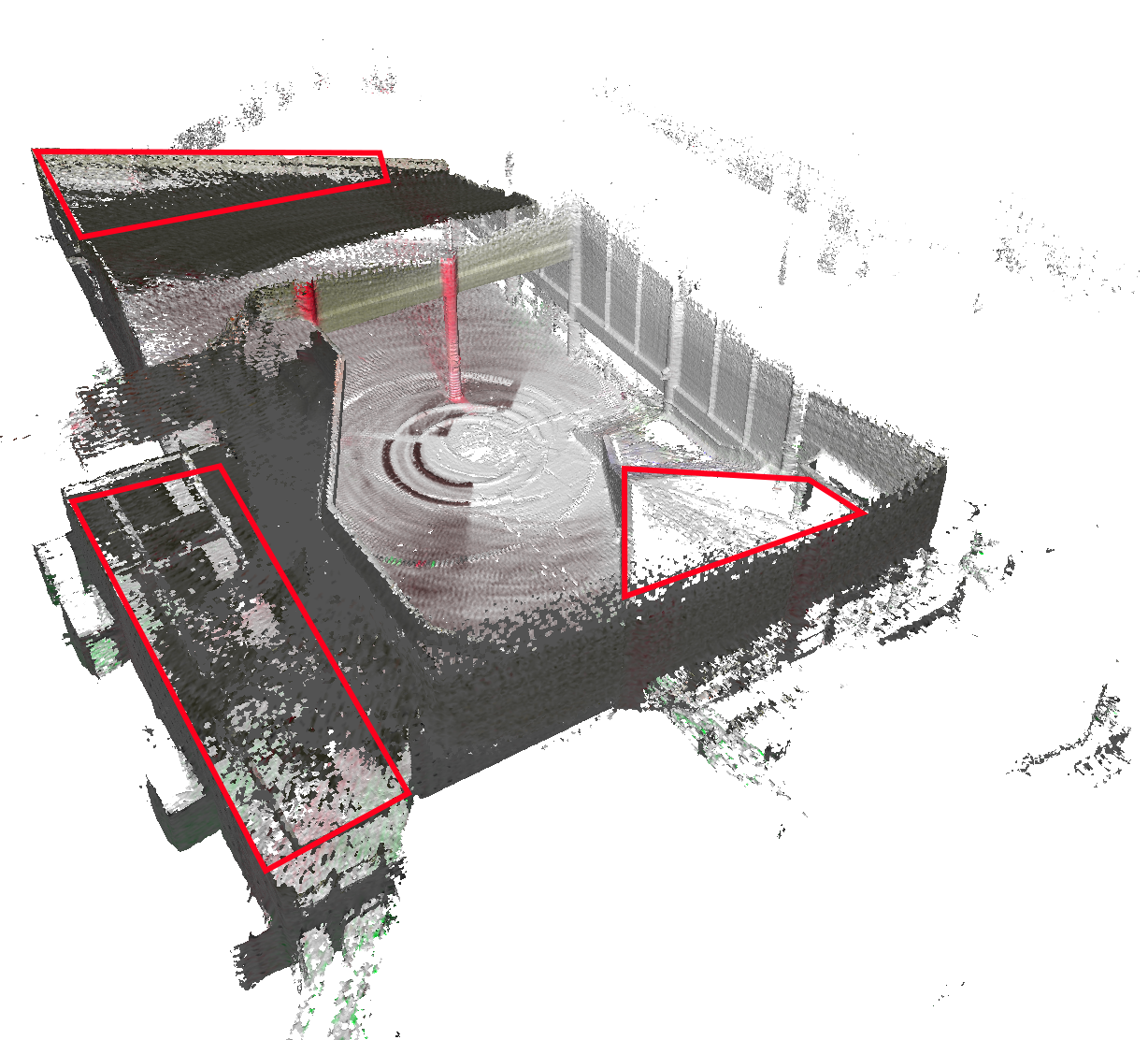}}
    \hspace{0.1in} % Optional: adds horizontal space between subfigures
    \subfigure[Ours]{\includegraphics[width=0.3\textwidth]{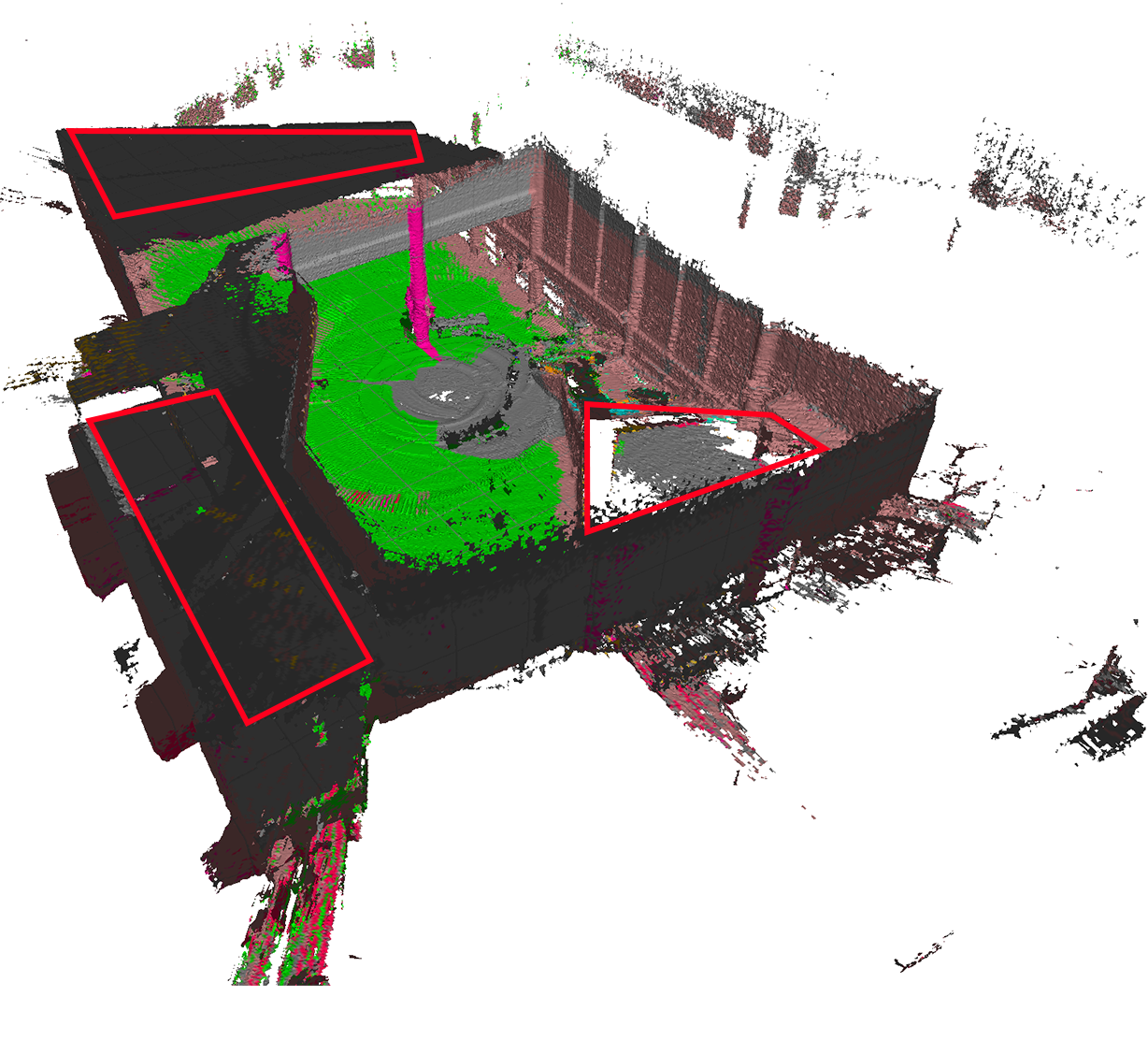}}
    \caption{3D reconstruction of NTU VIRAL (NYA01 Scene) dataset}
    \label{fig:qualitative comparison ntuviral}
\end{figure*}

In Figure~\ref{fig:qualitative comparison ntuviral}c, our reconstruction exhibits significantly higher surface completeness and fidelity in annotated areas without over-smoothing. Voxblox (Figure~\ref{fig:qualitative comparison ntuviral}b) fails to capture fine details due to static TSDF truncation and limited fusion updates, resulting in blurry and incomplete surfaces. ImMesh (Figure~\ref{fig:qualitative comparison ntuviral}a) shows sharp surfaces but lacks geometric continuity in occluded and sparse regions—noticeable on the floor, back wall, and annotated regions.

\subsection{Quantitative Results}\label{sec:Quantitative}
We evaluate on the Christ Church College scene (sequence~2) of the Oxford Spires dataset, selected because it contains an indoor cultural-heritage environment with terrestrial laser scanning (TLS) ground truth — aligning with our target application of large and complex indoor spaces. We adopt the accuracy, completeness, and F1 metrics commonly used 
in geometric reconstruction evaluation~\cite{Dai2017BundleFusion,immesh,zhu2023}. Each reconstructed mesh is sampled to 500{,}000 points and aligned to the ground truth via manual coarse registration followed by ICP 
refinement. The evaluation framework, including point sampling, 
outlier filtering, and metric definitions, is described below.

\newcommand{\R}{\mathcal{R}}   % reconstruction set
\newcommand{\G}{\mathcal{G}}   % ground-truth set
\newcommand{\norm}[1]{\left\lVert #1 \right\rVert_2}
\newcommand{\1}{\mathbf{1}}

\subsubsection{Setup.}
Let $\R = \{r_i\}_{i=1}^{N_\R}$ be the aligned reconstruction and
$\G = \{g_j\}_{j=1}^{N_\G}$ the ground truth. 
We define the nearest-neighbor distances as the minimum Euclidean distance
from each point in the reconstruction $\R$ to the closest point in the
ground truth $\G$, and vice versa:
\begin{align}
d_{\R\to\G}(r) &= \min_{g\in\G}\lVert r - g\rVert_2, &
d_{\G\to\R}(g) &= \min_{r\in\R}\lVert g - r\rVert_2.
\label{eq:nn}
\end{align}
To remove spurious correspondences we apply an outlier filter with radius
$D_{\max} > 0$. For each $r \in \R$ we keep its distance only if it lies within
this radius,
\begin{align}
\tilde d_{\R\to\G}(r) &=
\begin{cases}
d_{\R\to\G}(r), & d_{\R\to\G}(r)\le D_{\max},\\
\text{discard}, & \text{otherwise},
\end{cases}
\\[-0.25em]
\tilde d_{\G\to\R}(g) &=
\begin{cases}
d_{\G\to\R}(g), & d_{\G\to\R}(g)\le D_{\max},\\
\text{discard}, & \text{otherwise}.
\end{cases}
\end{align}
Let $\tilde\R,\tilde\G$ denote the \emph{radius-filtered subsets} of
$\R$ and $\G$, i.e., the points whose nearest-neighbor distance does not
exceed $D_{\max}$.

\subsubsection{Accuracy (RMSE \& percentage).} Accuracy reports the average geometric error of reconstruction points to the nearest ground-truth surface (lower is better).

Given a max-correspondence threshold $\tau$, we define the inlier set
$\mathcal{I}_\R(\tau)=\{\,r\in\tilde\R \mid d_{\R\to\G}(r)\le\tau\,\}$.
Throughout, we set $\tau = 0.30\,\mathrm{m}$, which is several times the expected range and registration noise of our indoor LiDAR setup (maximum range $\approx 150\,\mathrm{m}$), so larger deviations are treated as outliers rather than measurement noise. The (inlier) RMSE accuracy is
\begin{align}
\mathrm{Acc}^{\mathrm{RMSE}}_\tau
= \sqrt{\frac{1}{|\mathcal{I}_\R(\tau)|}\sum_{r\in\mathcal{I}_\R(\tau)} d_{\R\to\G}(r)^2}\,.
\end{align}
Complementarily, the \emph{percentage} (“precision at $\tau$”) reports the fraction of retained reconstruction points within $\tau$:
\begin{align}
\mathrm{Acc}^{\%}_\tau
= \frac{|\mathcal{I}_\R(\tau)|}{|\tilde\R|}\,.
\end{align}

\subsubsection{Completeness (RMSE \& percentage).} 
Completeness reports the average geometric miss from ground-truth points
to the nearest reconstructed surface (lower is better). Using the distances $d_{\G\to\R}(g)$ from \eqref{eq:nn} and the same
threshold $\tau$, define
$\mathcal{I}_\G(\tau)
= \{\,g\in\tilde\G \mid d_{\G\to\R}(g)\le\tau\,\}$.
The (inlier) RMSE completeness is
\begin{align}
\mathrm{Comp}^{\mathrm{RMSE}}_\tau
= \sqrt{\frac{1}{|\mathcal{I}_\G(\tau)|}\sum_{g\in\mathcal{I}_\G(\tau)} d_{\G\to\R}(g)^2}\,.
\end{align}
The corresponding \emph{percentage} (“recall at $\tau$”) is the fraction of retained ground-truth points within $\tau$:
\begin{align}
\mathrm{Comp}^{\%}_\tau
= \frac{|\mathcal{I}_\G(\tau)|}{|\tilde\G|}\,.
\end{align}

\subsubsection{F\textsubscript{1} score (percentage).}
A thresholded balance of precision and recall reported as a single percentage (higher is better).

For threshold $\tau$, we combine the percentage precision and recall via the harmonic mean:
\begin{align}
\mathrm{F1}_\tau
= \frac{2\,\mathrm{Acc}^{\%}_\tau\,\mathrm{Comp}^{\%}_\tau}
       {\mathrm{Acc}^{\%}_\tau+\mathrm{Comp}^{\%}_\tau}\,.
\end{align}

\begin{table}[htbp]
  \centering
  \begin{tabular}{|l|c|c|c|}\hline
    \textbf{Method} &
    \textbf{Acc. (cm) \down} &
    \textbf{Comp. (\%) \up} &
    \textbf{F1 (\%) \up} \\\hline
    ImMesh   & 17.40 & 95.99\% & 96.31\% \\\hline
    Voxblox  & 17.24 & 97.17\% & 96.98\% \\\hline
    Ours     & 14.54 & 98.55\% & 98.58\% \\\hline
  \end{tabular}
  \caption{Results from Oxford Spires dataset. Lower Acc. RMSE is better (\down); higher Completeness and F1 are better (\up).}
  \label{tab:geometric results}
\end{table}

% In Table~\ref{tab:geometric results}, we compare the geometric reconstruction quality quantitatively across three mesh pipelines (lower RMSE is better for Accuracy/Completeness; higher is better for F1). Our method achieves the lowest errors—14.54 cm accuracy and 14.39 cm completeness—improving over Voxblox (17.24/16.97 cm), and over ImMesh (17.40/18.40 cm). The simultaneous reduction in both accuracy and completeness errors indicates that our pipeline not only fits local surfaces more tightly but also covers a larger fraction of the ground-truth geometry. We attribute the improvement to our semantics-and uncertainty awareness in TSDF fusion handling, which preserve high-curvature structures while completing more planar surfaces, yielding cleaner, more faithful meshes. 
For conciseness, Table~\ref{tab:geometric results} reports the most informative metric per dimension: accuracy as RMSE (capturing the magnitude of surface error) and completeness as percentage (capturing coverage of the ground truth). In Table~\ref{tab:geometric results}, we compare the geometric reconstruction quality quantitatively across three mesh pipelines. Our method achieves the lowest accuracy error (14.54\,cm) and the highest completeness (98.55\%) and F1 (98.58\%), improving over Voxblox (17.24\,cm / 97.17\% / 96.98\%) and ImMesh (17.40\,cm / 95.99\% / 96.31\%). The simultaneous reduction in accuracy error and increase in completeness indicates that our pipeline not only fits local surfaces more tightly but also covers a larger fraction of the ground-truth geometry. We attribute the improvement to semantics- and uncertainty-aware TSDF fusion, which preserves high-curvature structures while completing planar surfaces, yielding cleaner, more faithful meshes.

\subsection{Uncertainty Results}
\label{sec:UncertainityResults}
\begin{table}[htbp]
    \centering
    \setlength{\tabcolsep}{5pt}
    \renewcommand{\arraystretch}{1.3} % mild stretch, combine with top rule

    \begin{tabular}{|l|c|c|c|c|}\hline
        \textbf{Dataset} &
        \textbf{\shortstack{\rule{0pt}{2.0ex}Cert.\\(Labelled)}} & 
        \textbf{\shortstack{\rule{0pt}{2.0ex}Sem.\\Unc.}} & 
        \textbf{\shortstack{\rule{0pt}{2.0ex}Geo.\\Unc.}} & 
        \textbf{\shortstack{\rule{0pt}{2.0ex}Both}} \\\hline
        NTU-VIRAL & 
        \shortstack{\rule{0pt}{2.0ex}4515082\\(83\%)} & 
        \shortstack{\rule{0pt}{2.0ex}776669\\(14\%)} & 
        \shortstack{\rule{0pt}{2.0ex}122051\\(2\%)} & 
        \shortstack{\rule{0pt}{2.0ex}33979\\(1\%)} \\\hline

        Oxford Spires & 
        \shortstack{\rule{0pt}{2.0ex}1348594\\(81\%)} & 
        \shortstack{\rule{0pt}{2.0ex}285470\\(17\%)} & 
        \shortstack{\rule{0pt}{2.0ex}23388\\(1\%)} & 
        \shortstack{\rule{0pt}{2.0ex}9364\\(1\%)} \\\hline
    \end{tabular}

    \caption{Proportion of samples under uncertainty categories.}
    \label{tab:uncertainty-breakdown}
\end{table}

\begin{figure*}[t]
    \centering
    \subfigure[Geometric Uncertainty (View:1)]{\includegraphics[width=0.4\textwidth]{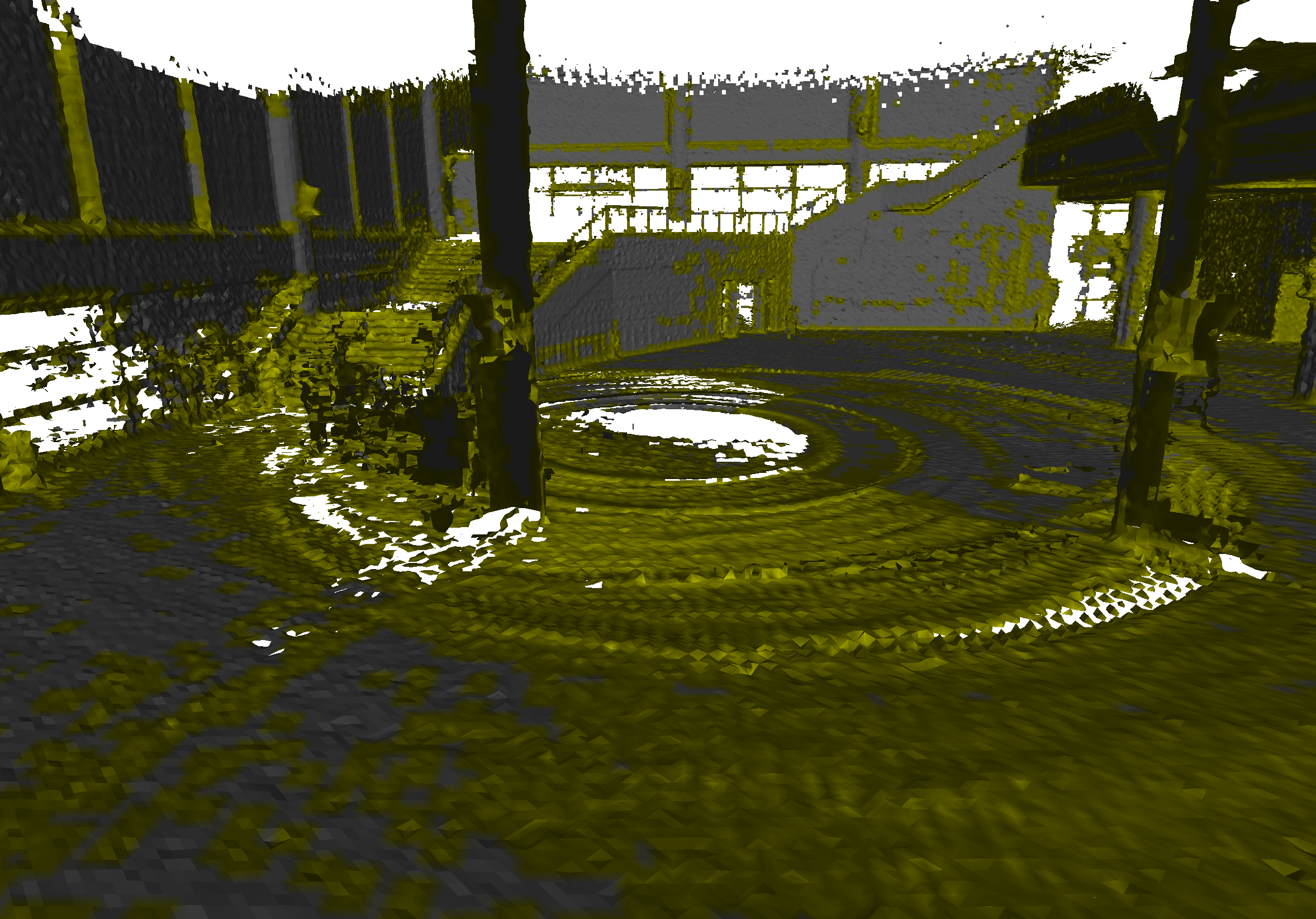}}
    \hspace{0.1in} % Optional: adds horizontal space between subfigures
    \subfigure[Geometric Uncertainty (View:2)]{\includegraphics[width=0.4\textwidth]{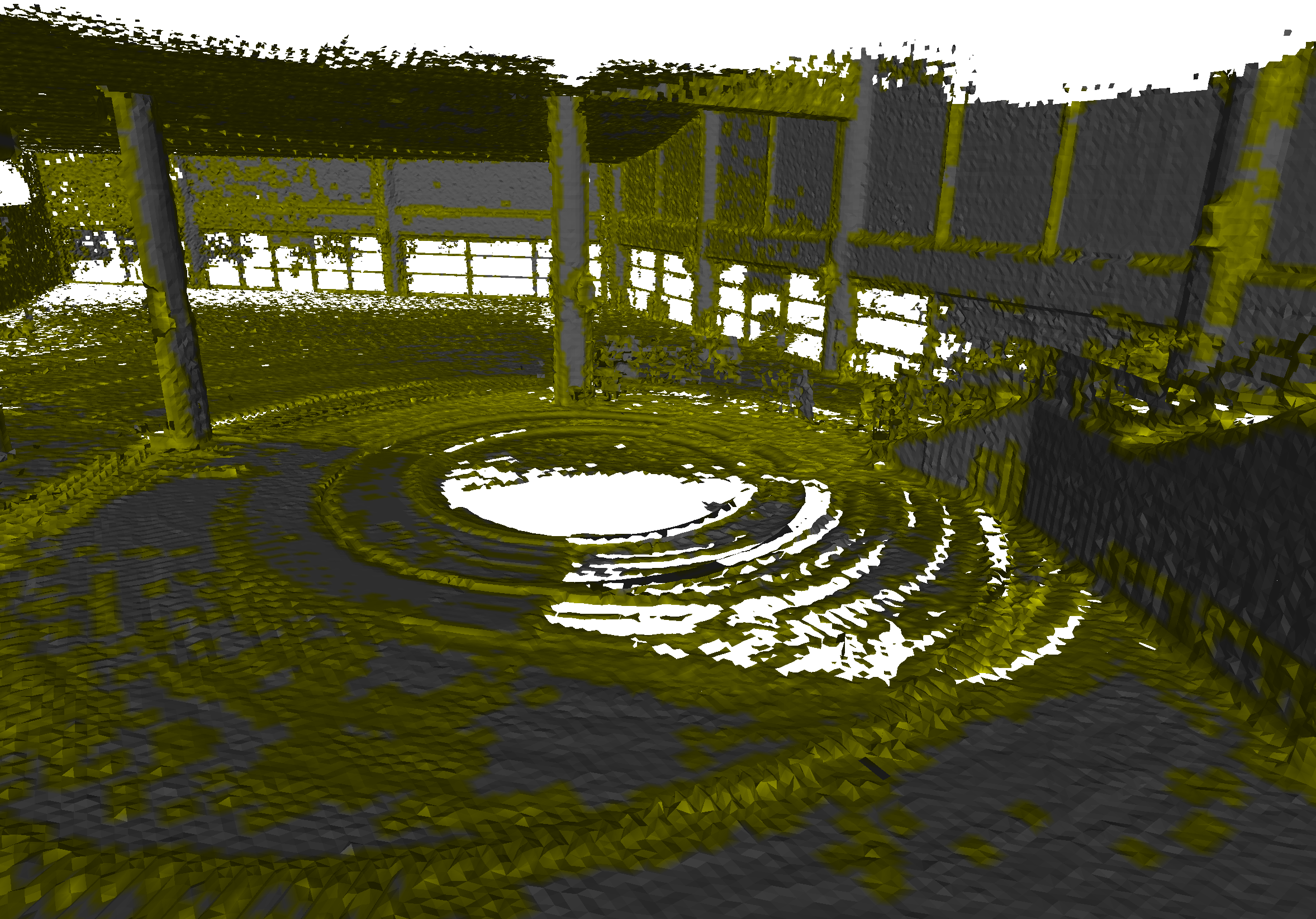}}
    \subfigure[Semantic Uncertainty (View:1)]{\includegraphics[width=0.4\textwidth]{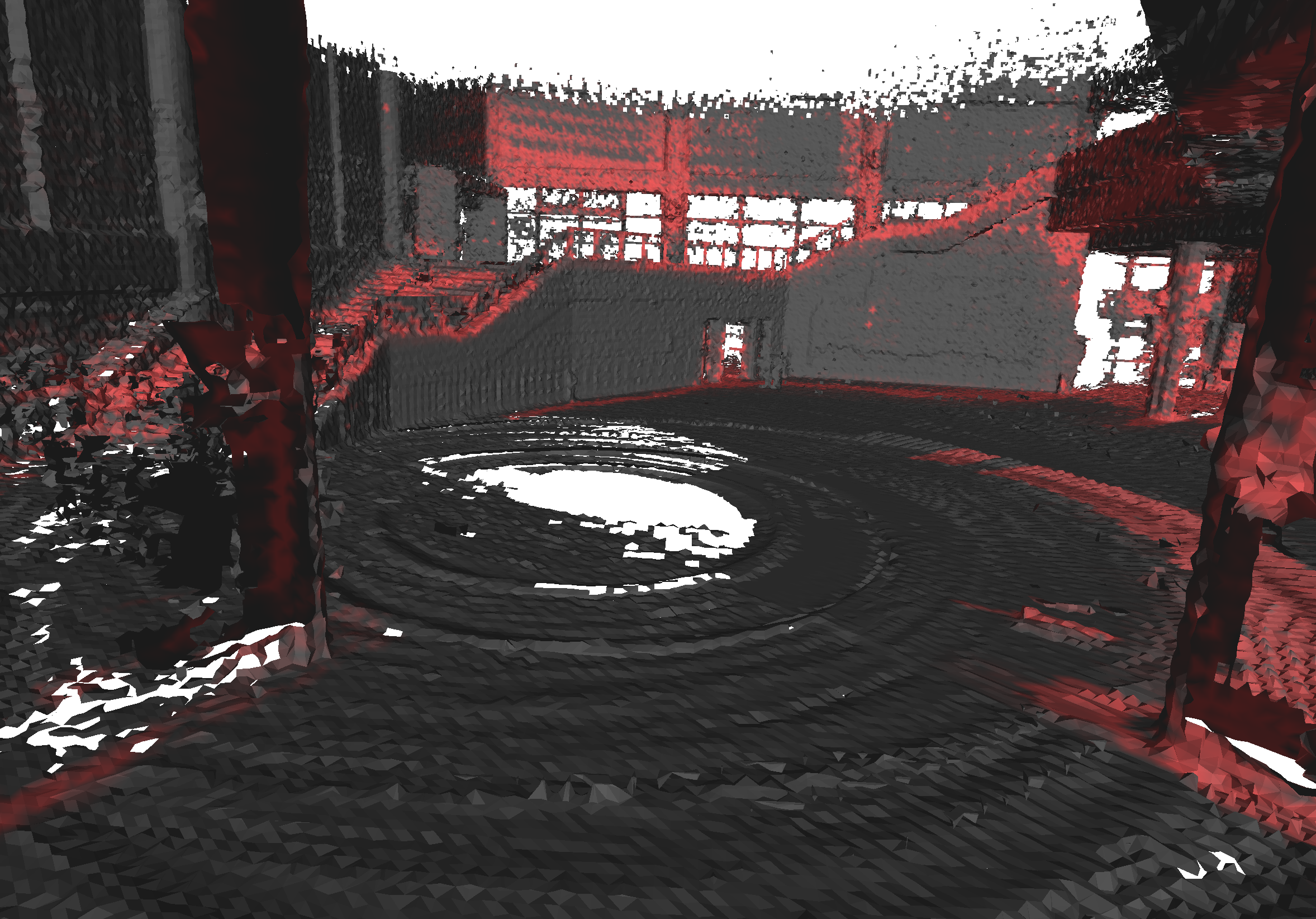}}
    \hspace{0.1in} % Optional: adds horizontal space between subfigures
    \subfigure[Semantic Uncertainty (View:2)]{\includegraphics[width=0.4\textwidth]{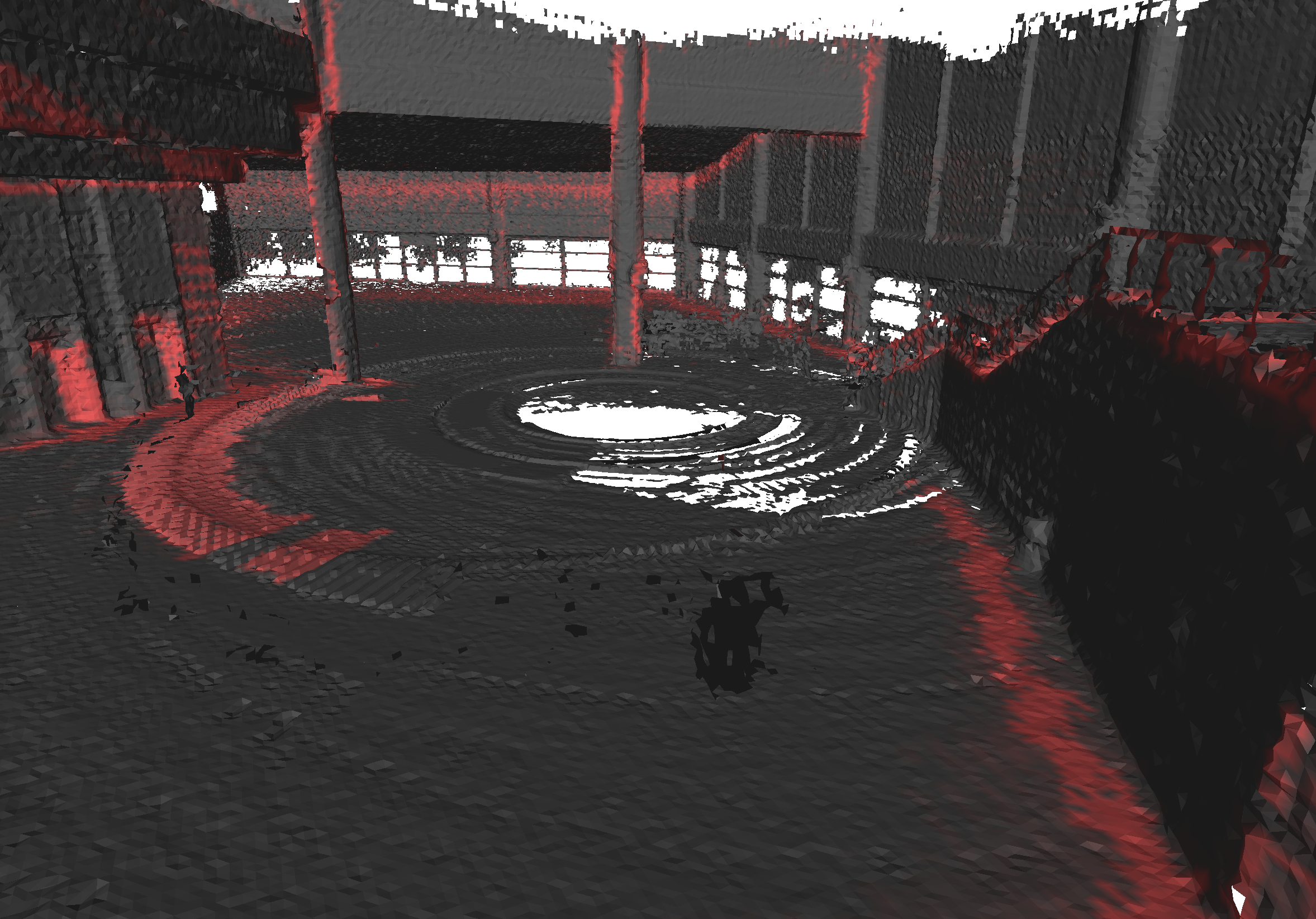}}  
    \caption{Analyzing the boundary uncertainty in 3D reconstruction of NTU VIRAL (NYA01) dataset}
    \label{fig:sem and geom uncertainity}
\end{figure*}

Table~\ref{tab:uncertainty-breakdown} and Figure~\ref{fig:sem and geom uncertainity} show a limited overlap between geometry and semantic uncertainties, indicating the presence of geometric-semantic inconsistencies. The semantic map (orange) registers class transitions even when geometry is smooth, while the geometric map (yellow) concentrates along depth discontinuities, occluding contours, and high-curvature creases. 

This divergent behavior arises from different error sources. Geometric cues capture minor mapping irregularities as boundaries due to inaccuracies in point cloud registration, and sensor noise. In contrast, semantic cues inherit inaccuracies from image segmentation, and projection artifacts. 
Modularity poses a coupling challenge in addition to data-level discrepancies. Time de-skewing (motion-induced distortions within a single LiDAR scan), surface priors, and voxel regularity assumptions could be refuted when integrating LIO back-ends or meshing libraries. These mismatches propagate through mesh decimation and regularization, reducing overall robustness.

\section{Conclusion}\label{sec:conclusion}

We explored incremental 3D mesh reconstruction by directly transferring VFM semantic segmentation labels onto LIO-registered LiDAR point clouds, followed by label-aware TSDF-based volumetric mapping. By incorporating class-conditioned truncation and confidence-weighted fusion, we improved geometric reconstruction quality in terms of both accuracy and completeness over geometry-only baselines. We further analysed the geometric-semantic inconsistencies inherent in the multi-modal TSDF framework, separately characterising their sources. The resulting mesh can be exported as a USD asset for downstream XR tooling. Limitations include odometry drift on longer trajectories, dependence on camera field-of-view coverage and calibration quality, and challenging indoor conditions such as thin structures, specular materials, and heavy clutter — which warrant further investigation.

\section{Acknowledgements}
\label{sec:acknowledgements}
This work was supported by the European Union under the Horizon Europe RIA grant agreement No. 101136006 (XTREME).
 
{
	\begin{spacing}{1.10}
		\normalsize
        
		\bibliography{biblography.bib} % Include your own bibliography (*.bib), style is given in isprs.cls
	\end{spacing}
}

\end{document}